\documentclass[conference]{IEEEtran}
\IEEEoverridecommandlockouts
\usepackage[nocompress]{cite}

\usepackage{amsmath,amssymb,amsfonts}
\usepackage{algorithmic}
\usepackage{graphicx}
\usepackage{textcomp}
\usepackage{xcolor}
\def\BibTeX{{\rm B\kern-.05em{\sc i\kern-.025em b}\kern-.08em
    T\kern-.1667em\lower.7ex\hbox{E}\kern-.125emX}}

\usepackage{comment}
\usepackage{subcaption}
\usepackage{algorithm}
\usepackage{algorithmic}
\usepackage{flushend}

\newcommand{\cmmnt}[1]{}

\def\BibTeX{{\rm B\kern-.05em{\sc i\kern-.025em b}\kern-.08em
    T\kern-.1667em\lower.7ex\hbox{E}\kern-.125emX}}

\begin{document}

\title{Efficient Low-Rank GNN Defense Against Structural Attacks}

 \author{\IEEEauthorblockN{Abdullah Alchihabi\qquad Qing En}
\IEEEauthorblockA{\textit{School of Computer Science, Carleton University} \\
Ottawa, Canada \\
abdullahalchihabi@cmail.carleton.ca, qingen@cunet.carleton.ca}
\and

\and
\IEEEauthorblockN{Yuhong Guo}
\IEEEauthorblockA{\textit{School of Computer Science, Carleton University}, Canada \\
\textit{Canada CIFAR AI Chair, Amii, Canada}\\
yuhong.guo@carleton.ca}
}

\maketitle

\begin{abstract}

Graph Neural Networks (GNNs) have been shown to possess strong representation abilities over graph data.
However, GNNs are vulnerable to adversarial attacks, 
and even minor perturbations to the graph structure can significantly degrade their performance. 
Existing methods either are ineffective against sophisticated attacks or require the optimization of dense adjacency matrices, which is time-consuming and prone to local minima. 
To remedy this problem, we propose an Efficient Low-Rank Graph Neural Network (ELR-GNN) defense method, 
which aims to learn low-rank and sparse graph structures for defending against adversarial attacks, 
ensuring effective defense with greater efficiency. 
	Specifically, ELR-GNN consists of two modules: a coarse low-rank estimation module and a fine-grained estimation module. 
The first module adopts the truncated Singular Value Decomposition (SVD) to initialize
	a low-rank estimate of the adjacency matrix, which serves as the starting point for optimizing the low-rank matrix. In the second module, the initial estimate is refined by jointly learning a low-rank sparse graph structure together with the GNN model. Sparsity is enforced on the learned low-rank adjacency matrix by pruning weak connections, which can reduce redundant data while maintaining valuable information. As a result, instead of using the dense adjacency matrix directly, ELR-GNN can learn a low-rank and sparse estimate of it in a simple, efficient, and easy to optimize manner. The experimental results demonstrate that ELR-GNN outperforms the state-of-the-art GNN defense methods in the literature, in addition to being very efficient and easy to train. 

\end{abstract}


\section{Introduction}

Graphs are ubiquitous data structures that can represent complex relationships between instances in various domains, 
such as social networks \cite{perozzi2014deepwalk}
and biological sciences \cite{zitnik2017predicting}.
Due to the widespread use of graphs, learning to effectively represent them 
is vital yet challenging.
Given their non-linear nature and capability of aggregating information from neighbouring nodes,
Graph Neural Networks (GNNs) have been widely adopted as a state-of-the-art architecture 
for learning with graph data and 
solving the node classification task, 
which is one fundamental and critical task in graph analysis. 

Despite their great performance, 
GNNs are vulnerable to adversarial attacks.
Small changes in the features of a few nodes or their corresponding connections 
in the graph may cause dramatic degradation in GNN performance \cite{dai2018adversarial,zugner2018adversarial},
This implies that imperceptible perturbations to the graph can significantly impact GNN performance.
In such cases, 
GNN models that lack robustness may present significant challenges to real-world privacy and security, 
particularly in  sectors like healthcare, communication networks, or finance.
Therefore, it is important to develop GNN models that are both efficient and resilient to state-of-the-art 
adversarial attacks.

Some works in the literature attempt to defend against adversarial attacks on graph structures by 
assigning larger weights to the edges connecting similar nodes and smaller weights to the edges connecting dissimilar nodes \cite{zhang2020gnnguard,zhu2019robust}.
Several contemporary methods pre-process the graph structures to satisfy certain desirable properties that are assumed to exist in clean graphs \cite{entezari2020all,wu2019}.
However, the aforementioned approaches only 
 address a relatively simple type of attack and
might confront significant challenges in defending against 
state-of-the-art global attacks
\cite{jin2020graph}.
In contrast, this work considers poisoning training-time attacks on graph structures, 
which represent the most formidable challenges to defend against \cite{zhu2021relationship}.
In this setting, the graph structure is tampered by 
some state-of-the-art attack methods, and  
the GNN model is trained on the poisoned graph structure, thereby introducing significant difficulties.

The fundamental challenges for developing effective methods 
to defend against adversarial attacks 
lie in the following two aspects.
(1) As GNNs employ message passing to propagate information across the graph,
attacks on local edges or nodes can be propagated to a large portion of the graph,
making it more difficult to defend against the 
state-of-the-art attacks.
(2) It is challenging to obtain GNN models that are both robust and efficient, 
as effective defense methods often 
require extensive computations over dense adjacency matrices.

In the meantime, it has been noted that 
challenging adversarial attacks (nettack, mettack) on a graph structure 
target the high-ranked (low-valued) singular components of its adjacency matrix, 
increasing the rank of the adjacency matrix 
while leaving the low-ranked singular components unperturbed \cite{entezari2020all,jin2020graph}.
Based on this observation, 
Entezari et al. recently introduced a simple pre-processing method for defending against adversarial attacks 
by obtaining a low-rank estimate of the adjacency matrix before training the GNN model \cite{entezari2020all}.
However, by separating the graph cleaning step and the GNN learning step,
this method leads to sub-optimal results
and lacks sufficient robustness against sophisticated global attacks.
In contrast, Pro-GNN \cite{jin2020graph} jointly learns a low-rank sparse graph structure with the GNN model.
Nevertheless, this approach still requires optimizing a dense adjacency matrix and minimizing its nuclear norm to 
guarantee the low-rank property, 
which involves difficult and computationally expensive optimization processes.

In this paper, we propose a novel Efficient Low-Rank Graph Neural Network (ELR-GNN) method, 
which is fast and efficient, 
for improving the robustness of GNNs in the face of adversarial attacks.
ELR-GNN focuses on learning a low-rank sparse estimate of the graph structure by estimating a low-rank adjacency matrix as the product of low-dimensional matrices instead of learning the dense adjacency matrix directly. 
Specifically, by computing the largest singular values and their corresponding singular vectors using the truncated Singular Value Decomposition (SVD), we first design a coarse low-rank estimation module to obtain elements mostly unaffected by the adversarial attacks, which provides a starting point to optimize the singular vector matrix. 
Next, we propose a fine-grained estimation module, where the low-rank estimate of the adjacency matrix is refined by jointly optimizing the singular vectors with the GNN model while keeping the obtained singular values fixed.
We incorporate sparsity into the learned low-rank adjacency matrix by pruning weak connections with low edge weights to remove redundant information while retaining important information. 
Additionally, we adopt the Frobenius norm to regularize the adjacency matrix 
for maintaining reasonable values. 
By combining weak edge pruning and Frobenius norm regularization, 
we can efficiently and quickly sparsify the adjacency matrix estimate.
Consequently, our proposed ELR-GNN can learn a low-rank and sparse estimate of the graph structure in an efficient and easy-to-optimize manner. 
Comprehensive experiments on various benchmark datasets and against multiple attack methods demonstrate 
that our proposed method is simple and fast, outperforming state-of-the-art defense methods.

\section{Related Works}

\subsection{Adversarial Attacks on GNNs}

Graph Neural Networks can be adversarially attacked through their node features, graph structure 
\cite{dai2018adversarial,xu2019} or both \cite{wu2019,zugner2018adversarial,zugner2019meta}.
These attacks on graphs can be grouped into two different types with different goals: non-targeted global attacks that aim to reduce the overall performance of GNNs and targeted attacks that aim to misclassify specific nodes in the graph. 
Meanwhile, adversarial attacks on GNNs can also be categorized based on the time of the attacks: 
poisoning attacks that take place prior to the training of the GNN models and 
evasion attacks that take place at test-time after the GNN model has been trained. 
Z\"{u}gner et al. proposed 
a targeted attack method, 
which iteratively perturbs the graph to maximize the degradation of the performance of a surrogate GNN model \cite{zugner2018adversarial}. 
Wu et al. introduced a method to attack both the graph structure and node features based on integrated gradients \cite{wu2019}. 
They also developed a defense approach based on these attack methods called GCN-Jaccard, where Jaccard similarity is used to pre-process the graph structure by deleting edges that connect nodes with similarity below some pre-defined threshold.
Z\"{u}gner and G\"{u}nnemann proposed a poisoning non-targeted adversarial attack method (mettack) to attack the graph structure based on meta-gradients where the graph structure is perturbed during training time by solving a min-max problem \cite{zugner2019meta}. 
Xu et al. proposed two gradient-based attack methods to attack the graph structure, named as projected gradient descent topology attack and min-max topology attack \cite{xu2019}.
  
However, not all attacks are equally effective. Wu et al. have shown that attacks on node features are significantly less effective in terms of degrading the GNNs performance than attacks on the graph structure \cite{wu2019}. Zhu et al. have also demonstrated that poisoning training-time attacks are more severe and harder to defend against compared to evasion test-time attacks \cite{zhu2021relationship}. Therefore, in this work, we focus on defending against poisoning training-time attacks on the graph structure (mettack and nettack) which are the most effective and challenging attacks to defend against.

\subsection{Adversarial Defense on GNNs}
Many approaches to defend GNNs against adversarial attacks have been proposed. 
Some works utilize pre-processing methods to filter the perturbed graph structure prior to the training stage 
\cite{entezari2020all,wu2019}. 
Other works use adversarial training to defend against adversarial attacks 
\cite{xu2019,zugner2019meta}. 
Zhu et al. proposed RGCN, in which the hidden node embeddings are represented as Gaussian distributions to absorb the effect of adversarial attacks in the covariance of the distribution and attention is used with the covariance matrix to aggregate messages from neighbouring nodes \cite{zhu2019robust}. 
Simp-GCN employs a novel adaptive message aggregation mechanism and self-supervised learning to preserve node similarities during GNN training \cite{jin2021node}. Elastic GNN (E-GNN) employs a novel $\ell_1$-based graph smoothing message aggregation function to improve the robustness of GNNs to adversarial attacks \cite{liu2021elastic}. 

In order to develop robust defense methods, the properties of the perturbed graph structures have also been investigated. 
Several works have shown that attack methods tend to connect dissimilar nodes more often than disconnecting similar nodes, as adding edges between dissimilar nodes hurts the performance more than deleting edges between similar nodes 
\cite{dai2018adversarial,wu2019,zugner2018adversarial}.
This has led to the development of homophily-based defense approaches 
that prune edges between dissimilar nodes \cite{wu2019,zhang2020gnnguard}. 
Zhang et al. introduced GNNGuard, which aims to connect similar nodes and disconnect dissimilar nodes using two components: 
a neighbour importance estimation component and a layer-wise graph memory component \cite{zhang2020gnnguard}.

Several works have demonstrated that adversarial attacks on the graph structure mainly affect the 
high-ranked (low-valued) singular components of the adjacency matrix, 
leaving the low-ranked singular components unaffected, thus causing an increase in the rank of the adjacency matrix \cite{entezari2020all,jin2020graph}.
This has inspired methods to defend against adversarial attacks by learning/estimating low-rank adjacency matrices to filter the impacts of the adversarial attacks on the graph structure. 
For instance, GCN-SVD pre-processes the input graph to obtain a low-rank estimate of the pre-perturbed graph structure using truncated SVD \cite{entezari2020all}.
However, due to the separation of the graph pre-processing step from the GNN training step, 
this approach lacks sufficient capacity in defending against sophisticated global attacks \cite{jin2020graph}. 
In contrast, Pro-GNN learns a clean graph structure jointly with the GNN model by optimizing the low-rank, sparsity and homophily properties of the estimated graph structure \cite{jin2020graph}. 
Nevertheless, Pro-GNN is computationally expensive as it requires minimizing the nuclear norm of the adjacency matrix at every training iteration.

%%%%%%%%%%%%%%
\begin{figure*}[t]
\centering

\includegraphics[width = 0.85 \textwidth,height=58mm]{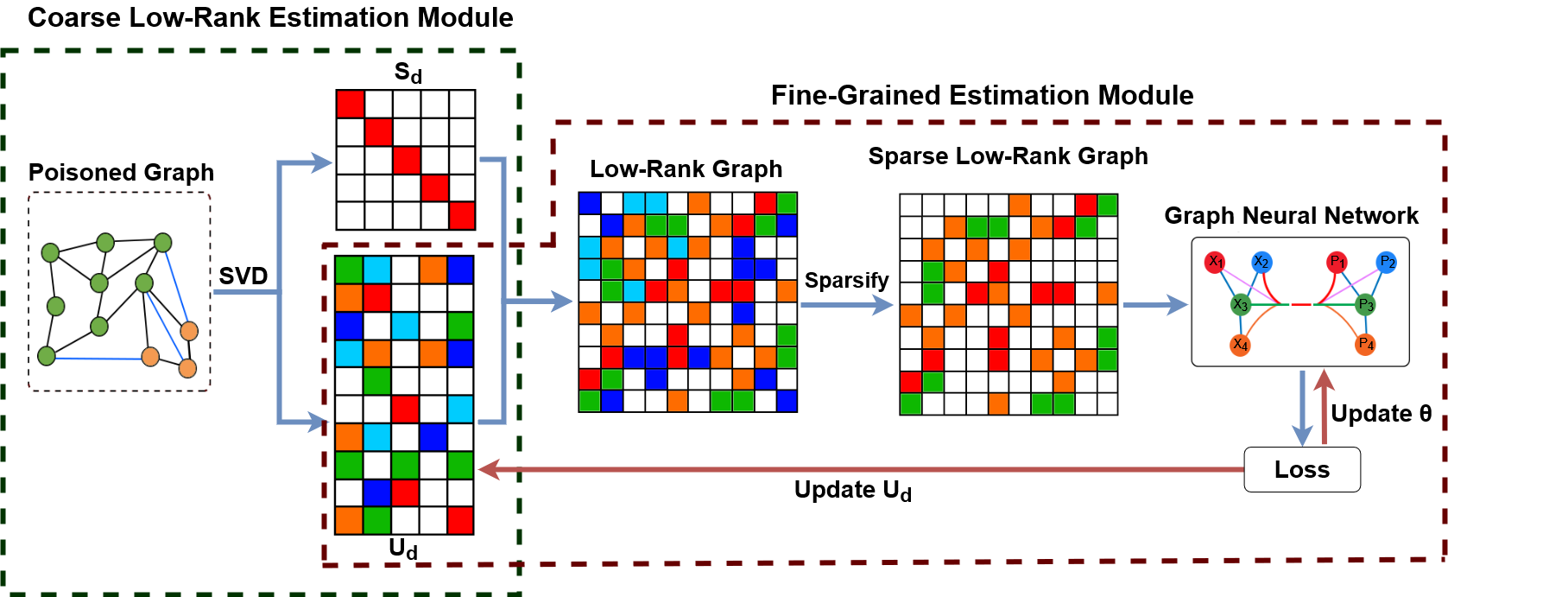} 

\caption{An illustration of the proposed ELR-GNN defense method. The framework is made up of two modules: the Coarse Low-Rank Estimation Module on the left side and the Fine-Grained Estimation Module on the right side.
}
\label{fig:diagram}
\end{figure*}
%%%%%%%%%%%%%%

\section{Method}
\subsection{Problem Setup}
We tackle the semi-supervised node classification task, 
which aims to predict the labels of unlabeled nodes
given a small number of labeled nodes.
We define the input data as a graph $G=(V,E)$, 
where $V$ is the set of nodes of the graph with $|V|=N$ and $E$ is the set of edges of the graph. 
Each node has a corresponding feature vector of size $D$, and the feature vectors of all the nodes 
in the graph form a matrix $X\in\mathbb{R}^{N\times D}$. 
The set of edges $E$ is represented as an adjacency matrix $A$ of size $N \times N$, 
which we assume to be symmetric (i.e., an undirected graph), 
with real-valued weights or binary values. 
In addition, 
the adjacency matrix $A$ can be either clean or 
perturbed by an adversarial attack method prior to the training of the GNN (i.e. poisoning attack). 
The nodes in the graph $V$ are split into two subsets: the set of labeled nodes $V_{\ell}$ and the set of unlabeled nodes $V_u$. Each labeled node has a corresponding label indicator vector of size $C$, where $C$ is the number of classes. The label vectors of all the labeled nodes constitute a label indicator matrix $Y^{\ell} \in \{0,1\}^{N_{\ell}\times C}$, 
where $N_{\ell}$ is the number of labeled nodes in the graph.

A GNN can be deployed on the graph data 
to predict the node labels as follows:
\begin{equation}
    P = f_{\Theta}(X,A),   
\end{equation}
where $f$ denotes the prediction function produced by the GNN parametrized with $\Theta$,
which takes the initial node features $X$ and adjacency matrix $A$ as input and 
outputs the class prediction probabilities $P$ over all the nodes. 
Typically, the GNN is trained to minimize the classification loss 
on the labeled nodes (cross-entropy loss): 
\begin{equation}
	\mathcal{L}_{CE} =  \sum_{i \in V_{\ell}} \ell_{CE}(P_i,Y^\ell_i),
\end{equation}
where $\ell_{CE}$ denotes the cross-entropy loss function, $P_i$, and $Y^\ell_i$ denote the predicted class probability vector and true label indicator vector for node $i$ respectively.

\subsection{The Proposed Method}

In this section, we present the proposed method ELR-GNN, which aims to defend GNNs against poisoning structural attacks. 
ELR-GNN learns a low-rank sparse estimate of the adjacency matrix as the product of 
two low-dimensional matrices: 
the low-dimensional singular value matrix and the low-dimensional singular vector matrix.

The overall architecture of ELR-GNN is illustrated in Figure \ref{fig:diagram}.
It has two modules: a coarse low-rank estimation module and a fine-grained estimation module. 
In the coarse low-rank estimation module,
we utilize the truncated Singular Value Decomposition (SVD) \cite{halko2011finding}  
to calculate the largest $d$ singular values of the adjacency matrix and their corresponding singular vectors. 
Truncated SVD has the nice property of being able to scale up with large sparse datasets. 
The low-dimensional singular value matrix is then formed with these largest $d$ singular values, 
while the low-dimensional singular vector matrix is initialized with the corresponding singular vectors. 
In this manner, we acquire a starting point for optimizing a low-dimensional singular vector matrix, 
which will be used with the singular value matrix to estimate a low-rank adjacency matrix efficiently. 
In the fine-grained estimation module,  
we jointly learn the low-dimensional singular vector matrix with the GNN model 
to further improve the low-rank estimate of the adjacency matrix.
In order to improve the robustness and reduce noise, 
we promote sparsity within the learned adjacency matrix  
by deleting the weak connections with weights below a pre-defined threshold.
In doing so, we anticipate obtaining
a robust sparse low-rank estimate of the adjacency matrix 
that are effective against sophisticated adversarial attacks.
Below, we elaborate on the two modules.

\subsubsection{Coarse Low-Rank Estimation Module}

The coarse low-rank estimation module aims to compute the singular value matrix and initialize 
the low-dimensional singular vector matrix, 
which allows us to estimate a coarse low-rank adjacency matrix from the perturbed adjacency matrix. 

To achieve that, singular value decomposition (SVD) can be deployed to decompose the adjacency matrix $A$ as follows: 
\begin{equation}
      A = U S V^\top,
\end{equation}
where $S\in\mathbb{R}^{N \times N}$ is the diagonal matrix of the singular values of $A$,
and $U, V \in\mathbb{R}^{N\times N}$ are orthogonal matrices whose columns are the left singular vectors and right singular vectors, respectively.
For undirected graphs with symmetric adjacency matrices, we have $U=V$. 
Meanwhile, it has been demonstrated in previous works
that adversarial attacks on the graph structure are typically high-rank attacks
\cite{entezari2020all,jin2020graph},
which means 
these attacks mainly affect the singular components
corresponding to lower singular values while leaving the ones associated with higher singular values unaffected.
As such, 
we only need the largest $d$ singular values and their corresponding singular vectors 
for our low-rank estimation of the adjacency matrix,
aiming to mitigate the impact of the attacks. 
Specifically, we define $S_d\in \mathbb{R}^{d \times d}$ and $U_d\in \mathbb{R}^{N \times d}$ 
as the singular value matrix and low-dimensional singular vector matrix, respectively, that 
correspond to the largest $d$ singular values of the adjacency matrix $A$.
That is, $S_d$ is a diagonal matrix of the largest $d$ singular values of $A$, 
and $U_d$ contains the corresponding singular vectors as its columns, 
where $d$ is a predefined hyper-parameter that is selected using cross-validation. 
Moreover, $S_d$ and $U_d$ can actually be obtained using an efficient truncated SVD algorithm
\cite{halko2011finding}
without full singular value decomposition. 

The singular value matrix $S_d$ and the singular vector matrix $U_d$ will then 
be used to estimate a low-rank adjacency matrix.
In particular, we construct a $d$-rank estimate of $A$ by using $U_d$ and $S_d$ as follows:
\begin{equation}
    A_{d} =  \Lambda \Lambda^{\top}, \quad  \text{where} \;  \Lambda = U_{d} S^{1/2}_{d}.
\end{equation}
This pre-training estimate however can only serve as an initial point for effectively 
defending against sophisticated adversarial structural attacks \cite{jin2020graph}. 
We will further improve the estimate using a fine-grained estimation module 
by jointly learning the low-dimensional matrix $U_d$ and the GNN model.

%%%%%%%%%%%%%%%%
\begin{table*}[!t] 
\caption{Mean classification accuracy (standard deviation is within brackets) on Cora (top part), CiteSeer (middle part) and PolBlogs (bottom part) under global non-targeted training-time attack (mettack) with different perturbation rates (0\%, 5\%, 10\%, 15\%, 20\%, 25\%).
}

\setlength{\tabcolsep}{4pt}
\resizebox{.99\textwidth}{!}
	{
\begin{tabular}{l|l|l|l|l|l|l|l|l|l|l|l}
	\hline
         & Ptb Rate & GCN  & GAT    & RGCN   & GCN-Jacc & GCN-SVD & Pro-GNN-fs  & Pro-GNN & SimP-GCN & E-GNN &  ELR-GNN        \\
\hline
Cora     & 0\%             & ${83.5}_{(0.4)}$ & ${83.9}_{(0.6)}$ & ${83.0}_{(0.4)}$ & ${82.0}_{(0.5)}$   
         & ${80.6}_{(0.4)}$        & ${83.4}_{(0.5)}$     & ${82.9}_{(0.2)}$ & $82.7_{(0.3)}$ & $\mathbf{85.8}_{(0.4)}$  &  ${80.7}_{(0.5)}$ \\
         & 5\%             & ${76.5}_{(0.7)}$          & ${80.4}_{(0.7)}$ & ${77.4}_{(0.3)}$ & ${79.1}_{(0.5)}$      & ${78.3}_{(0.5)}$        & $\mathbf{82.7}_{(0.3)}$     & ${82.2}_{(0.4)}$ & $78.9_{(0.9)}$ &$82.2_{(0.9)}$ &  ${80.5}_{(0.3)}$ \\
         & 10\%            & ${70.3}_{(1.2)}$ & ${75.6}_{(0.5)}$  & ${72.2}_{(0.3)}$    & ${75.1}_{(0.7)}$          
         & ${71.4}_{(0.8)}$           & ${77.9}_{(0.8)}$     & ${79.0}_{(0.5)}$  & $76.5_{(1.0)}$ &	$78.8_{(1.7)}$ &  $\mathbf{79.1}_{(0.6)}$ \\
         & 15\%            & ${65.1}_{(0.7)}$          & ${69.7}_{(1.2)}$         & ${66.8}_{(0.3)}$    & ${71.0}_{(0.6)}$         & ${66.6}_{(1.1)}$          & ${76.0}_{(1.1)}$              & ${76.4}_{(1.2)}$  
         &    $74.5_{(2.3)}$	  &  $77.2_{(1.6)}$       &  $\mathbf{77.6}_{(1.4)}$ \\
         & 20\%          & ${59.5}_{(2.7)}$          & ${59.9}_{(0.9)}$   & ${59.2}_{(0.3)}$          & ${65.7}_{(0.8)}$         
         & ${58.9}_{(1.1)}$           & ${68.7}_{(5.8)}$   & ${73.3}_{(1.5)}$  
         &     $73.0_{(2.9)}$      &	$70.5_{(1.3)}$          &  $\mathbf{77.7}_{(0.7)}$ \\
         & 25\%            & $47.5_{(1.9)}$          & ${54.7}_{(0.7)}$          & ${50.5}_{(0.7)}$         & ${60.8}_{(1.0)}$     & ${52.0}_{(1.1)}$           & ${56.5}_{(2.5)}$      & ${69.7}_{(1.6)}$   
         &  $70.5_{(3.2)}$ & \multicolumn{1}{|c|}{$-$} & $\mathbf{76.7}_{(0.7)}$ \\
\hline
Citeseer & 0\%            & ${71.9}_{(0.5)}$ & ${73.2}_{(0.8)}$  & ${71.2}_{(0.8)}$          & ${72.1}_{(0.6)}$      
          & ${70.6}_{(0.3)}$           & ${73.2}_{(0.3)}$     & ${73.2}_{(0.6)}$ 
          &  $\mathbf{73.9}_{(0.4)}$  &    ${73.8}_{(0.6)}$     &  ${73.4}_{(0.4)}$ \\
         & 5\%             & ${70.8}_{(0.6)}$          & ${72.8}_{(0.8)}$ & ${70.5}_{(0.4)}$         & ${70.5}_{(0.9)}$            & ${68.8}_{(0.7)}$    & ${73.0}_{(0.3)}$    & ${72.9}_{(0.5)}$  
          &  ${73.8}_{(0.3)}$      &   ${72.9}_{(0.5)}$          & $\mathbf{74.2}_{(0.5)}$ \\
         & 10\%            & ${67.5}_{(0.8)}$          & ${70.6}_{(0.4)}$          & ${67.7}_{(0.3)}$  & ${69.5}_{(0.5)}$          & ${68.8}_{(0.6)}$     & ${72.4}_{(0.5)}$    & ${72.5}_{(0.7)}$ 
         & ${71.5}_{(1.0)}$  & ${72.6}_{(0.4)}$  & $\mathbf{73.2}_{(0.8)}$ \\
         & 15\%            & ${64.5}_{(1.1)}$          & ${69.0}_{(1.0)}$   & ${65.6}_{(0.3)}$   & ${65.9}_{(0.9)}$               & ${63.2}_{(0.9)}$  & ${70.8}_{(0.8)}$      & ${72.0}_{(1.1)}$  
         & ${70.7}_{(1.3)}$      & ${71.9}_{(0.7)}$     & $\mathbf{74.6}_{(0.3)}$ \\
         & 20\%            & ${62.0}_{(3.4)}$          & ${61.0}_{(1.5)}$  & ${62.4}_{(1.2)}$  & ${59.3}_{(1.4)}$ 
         & ${58.5}_{(1.0)}$  & ${66.1}_{(2.3)}$      & ${70.0}_{(2.2)}$ & ${67.4}_{(1.4)}$ & 	${64.7}_{(0.8)}$ 
           & $\mathbf{71.6}_{(1.0)}$ \\
         & 25\%            & ${56.9}_{(2.0)}$          & ${61.8}_{(1.1)}$  & ${55.3}_{(0.6)}$  & ${59.8}_{(1.4)}$ 
         & ${57.1}_{(1.8)}$  & ${66.1}_{(2.3)}$      & ${68.9}_{(2.7)}$  & ${67.3}_{(1.7)}$       &    \multicolumn{1}{|c|}{$-$}       & $\mathbf{73.2}_{(0.6)}$ \\
\hline
Polblogs & 0\%      & ${95.6}_{(0.3)}$ & ${95.3}_{(0.2)}$ & ${95.2}_{(0.1)}$ & \multicolumn{1}{|c|}{$-$} 
        & ${95.3}_{(0.1)}$  & ${93.2}_{(0.6)}$     &  \multicolumn{1}{|c|}{$-$} &  ${95.1}_{(0.3)}$ & ${95.8}_{(0.3)}$  & $\mathbf{95.9}_{(0.1)}$ \\
         & 5\%             & ${73.0}_{(0.8)}$          & ${83.6}_{(1.4)}$    & ${74.3}_{(0.1)}$    & \multicolumn{1}{|c|}{$-$}     & ${89.0}_{(0.2)}$       & ${93.2}_{(0.1)}$     & \multicolumn{1}{|c|}{$-$}       
        &  ${72.8}_{(0.2)}$    &  ${83.0}_{(0.3)}$    & $\mathbf{94.5}_{(0.2)}$ \\
         & 10\%    & ${70.7}_{(1.1)}$    & ${76.3}_{(0.8)}$    & ${71.0}_{(0.3)}$     & \multicolumn{1}{|c|}{$-$}                  & ${81.2}_{(0.4)}$    & $\mathbf{89.4}_{(1.0)}$     & \multicolumn{1}{|c|}{$-$}       
        &  ${72.9}_{(0.6)}$ &   ${81.6}_{(0.3)}$ & ${88.2}_{(0.1)}$ \\
         & 15\%            & ${64.9}_{(1.9)}$       & ${68.8}_{(1.1)}$  & ${67.2}_{(0.3)}$   & \multicolumn{1}{|c|}{$-$}           & ${68.1}_{(3.7)}$    & $\mathbf{86.0}_{(2.2)}$     & \multicolumn{1}{|c|}{$-$}   
         & ${50.8}_{(1.3)}$  &  ${78.7}_{(0.5)}$  & ${80.4}_{(0.54)}$ \\
         & 20\%    & ${51.2}_{(1.2)}$  & ${51.5}_{(1.6)}$ & ${59.8}_{(0.3)}$     & \multicolumn{1}{|c|}{$-$}                       & ${57.3}_{(3.1)}$      & $\mathbf{79.5}_{(5.6)}$     & \multicolumn{1}{|c|}{$-$}   
         &    ${49.0}_{(1.5)}$   &   ${77.5}_{(0.2)}$ & ${77.5}_{(2.0)}$          \\
         & 25\%    & ${49.2}_{(1.3)}$     & ${51.1}_{(1.4)}$   & ${56.0}_{(0.5)}$    & \multicolumn{1}{|c|}{$-$}                   & ${48.6}_{(9.9)}$   & ${63.1}_{(4.4)}$  & \multicolumn{1}{|c|}{$-$}   & $48.2_{(1.8)}$ &   \multicolumn{1}{|c|}{$-$}  & $\mathbf{76.7}_{(1.2)}$ \\ 
         \hline
\end{tabular}}
\label{table:meta}

\end{table*}
%%%%%%%%%%%%%%%%

\subsubsection{Fine-Grained Estimation Module}
The fine-grained estimation module aims to learn a general low-dimensional matrix $U_d$
during GNN training
to approximate the singular vector matrix
and provide a fine-grained low-rank estimation
for the adjacency matrix $A$. 

Specifically, during the joint training stage 
we have $S_d$ fixed since the inconspicuous adversarial structural attacks do not alter the larger singular values of the adjacency matrix. 
We only update the GNN parameters $\Theta$ and the low-rank matrix $U_d$, which consequently leads to the update 
on the low-rank estimate of the adjacency matrix and affects GNN. 

Moreover, motivated by the fact that most real-world graphs are sparse in addition to being low rank, 
we further propose to sparsify the low-rank estimate of the adjacency matrix,  
which can significantly reduce computational overhead \cite{jin2020graph,zhou2013learning}.
To this end, 
we delete the weak connections in the estimated adjacency matrix, 
whose weights are smaller than a pre-defined threshold $\epsilon$: 
\begin{equation}
    A_{d}(i,j)= \begin{cases}
    A_{d}(i,j),& \text{if } \, A_{d}(i,j) \geq \epsilon \\
    0,              & \text{otherwise}
\end{cases}
\end{equation}
where the threshold  hyper-parameter $\epsilon$ 
controls the sparsity level of our low-rank estimate and 
is determined using cross-validation.
Following the sparsification of $A_{d}$, we further normalize this estimated sparse and low-rank adjacency matrix as follows: 
\begin{equation}
        \tilde{A}_{d}= D^{-\frac{1}{2}}A_{d} D^{-\frac{1}{2}},
\end{equation}
where $D$ is the diagonal degree matrix computed from $A_{d}$, such that $D_{ii}=\sum_j A_{ij}$.
This normalized adjacency matrix $\tilde{A}_{d}$ is then used as input for 
the GNN model.

As the adversarial attacks typically only perturb a minimal number of edges 
to degrade the performance of the GNN model, 
it is reasonable to make the learned sparse low-rank estimation of the adjacency matrix,
$\tilde{A}_{d}$,
to be similar to the original input adjacency matrix, $A$.  
This can be achieved by deploying the following similarity regularization 
term based on the Frobenius distance when learning the matrix $U_d$:
\begin{equation}
	\mathcal{L}_{\scriptsize Sim} =  \| A -  \tilde{A}_{d} \|^{2}_F,
\end{equation}
where $\|.\|_F$ denotes the Frobenius norm. 
Moreover, we also utilize a Frobenius norm regularization term over 
$\Lambda=U_dS_d^{1/2}$
during learning:
\begin{equation}
	\mathcal{L}_{\scriptsize Fr} =  \| \Lambda \|^{2}_F.
\end{equation}
This regularization term works in tandem with the pruning of weak edges to 
regularize the statistical distributivity of the adjacency matrix in an efficient manner.

In the end, we use the following overall loss function 
to jointly learn $U_d$ and $\Theta$ for the proposed ELR-GNN:
\begin{equation}
	\min_{\Theta, U_d }\; 
	\mathcal{L}   = \mathcal{L}_{CE}  +   \lambda_{sim}\,  \mathcal{L}_{Sim} + \lambda_{Fr}\,    \mathcal{L}_{Fr}.
\end{equation}
We solve this joint minimization problem using a simple alternating optimization procedure,
which updates one variable matrix while keeping the other one fixed. 
Specifically, when updating $\Theta$ for GNN learning,  
we keep the current $U_d$ fixed and only minimize
the cross-entropy loss:
 \begin{equation}
	\min_{\Theta }\; \mathcal{L}_{CE}.
\end{equation}
When updating $U_d$, we hold $\Theta$ fixed and minimize the overall loss function: 
\begin{equation}
	\min_{ U_d }\; \mathcal{L}   = \mathcal{L}_{CE}  +    \lambda_{sim}\,  \mathcal{L}_{Sim} + \lambda_{Fr}\,    \mathcal{L}_{Fr}
\end{equation}

\section{Experiments}
We evaluated the proposed method against training time structural attacks and 
conducted experiments under non-targeted global attacks (mettack), targeted attacks (nettacks) and random attacks.

\subsection{Experiment Settings}

\subsubsection{Datasets \& Baselines}
Three challenging datasets are used to evaluate our proposed ELR-GNN: two citation datasets (Cora, CiteSeer) \cite{sen2008collective} 
and one blog dataset (PolBlogs) \cite{adamic2005political}. 
For all the three datasets, we utilized the same train/validation/test node split adopted by Jin et al. \cite{jin2020graph}, where 10\% of the nodes are randomly assigned to the labeled train set, 10\% of the nodes are randomly assigned to the validation set, and the remaining 80\% of the nodes are assigned to the test set. 
In the case of the PolBlogs dataset, nodes are not associated with any features; 
therefore, we used an identity matrix as the nodes feature matrix. 
We compared the proposed ELR-GNN method with the following baselines:
Graph Convolution Networks (GCN) \cite{kipf2016semi}, Graph Attention Networks (GAT) \cite{velivckovic2017graph}, Robust Graph Convolution Networks (RGCN) \cite{zhu2019robust}, GCN-Jaccard \cite{wu2019}, GCN-SVD \cite{entezari2020all}, Pro-GNN \cite{jin2020graph}, Pro-GNN-fs \cite{jin2020graph}, Simp-GCN \cite{jin2021node} and Elastic-GNN (E-GNN) \cite{liu2021elastic}.
Among these baselines, GCN and GAT are plain GNN models, while the others have all adopted some defence strategies or methods.

%%%%%%%%%%%%%%%%
\begin{figure*}[t!]
\centering
\includegraphics[width=0.69\textwidth]{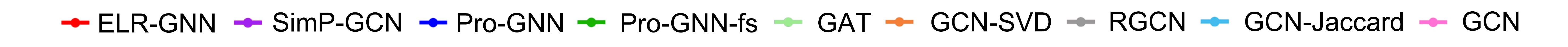}

\begin{subfigure}{0.28\textwidth}
\centering
\includegraphics[width = \textwidth,height = 1.7in]{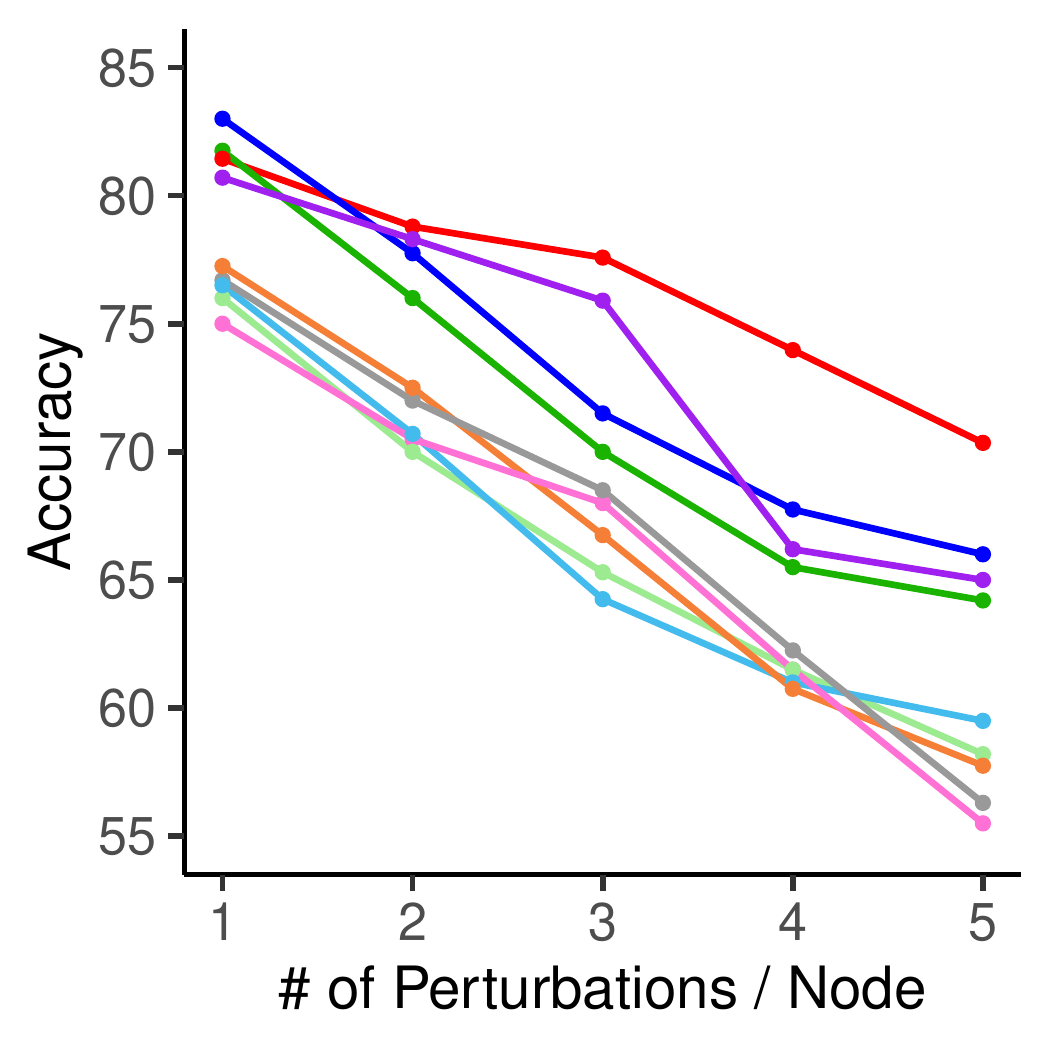} 
\caption{Cora}
\end{subfigure}
\begin{subfigure}{0.28\textwidth}
\centering
\includegraphics[width = \textwidth,height=1.7in]{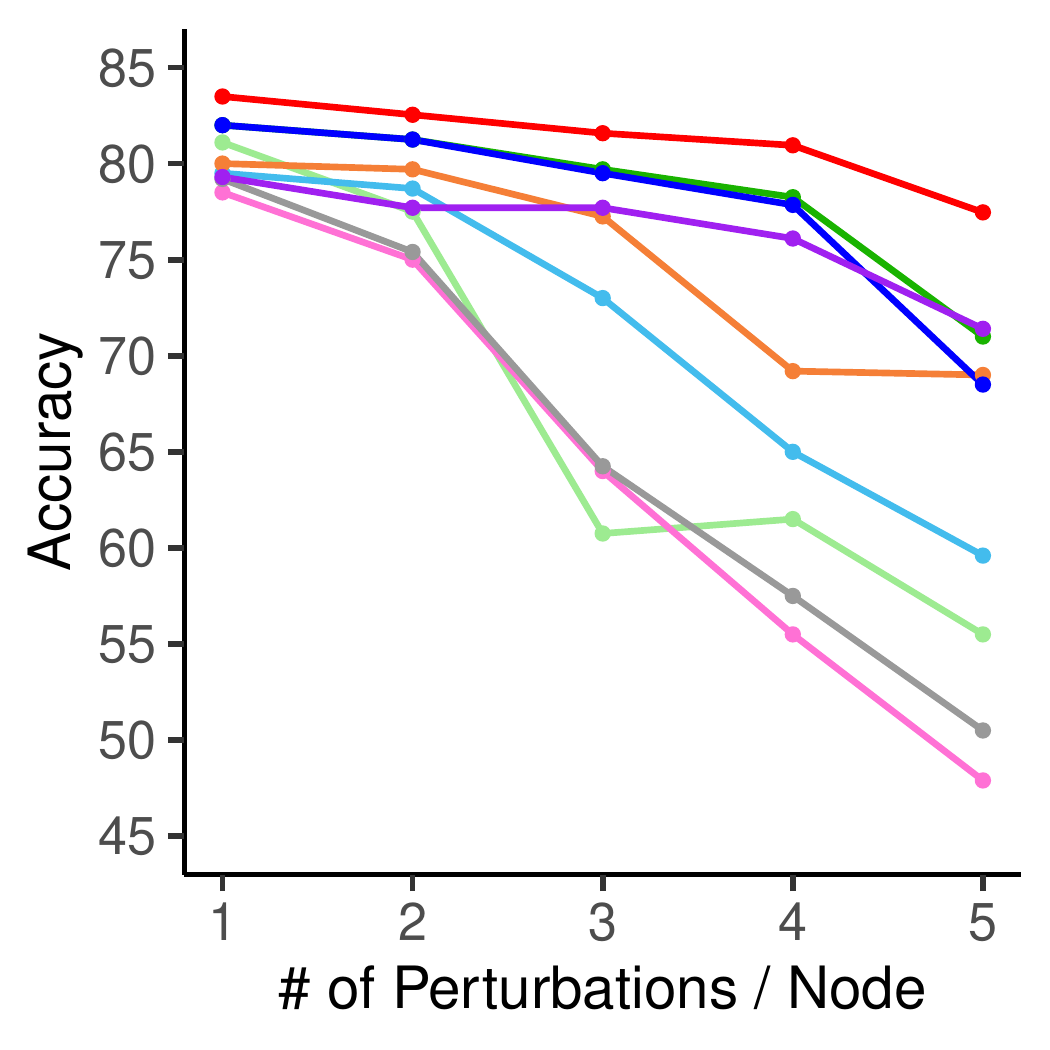}
\caption{CiteSeer}
\end{subfigure}
\begin{subfigure}{0.28\textwidth}
\centering
\includegraphics[width = \textwidth,height=1.7in]{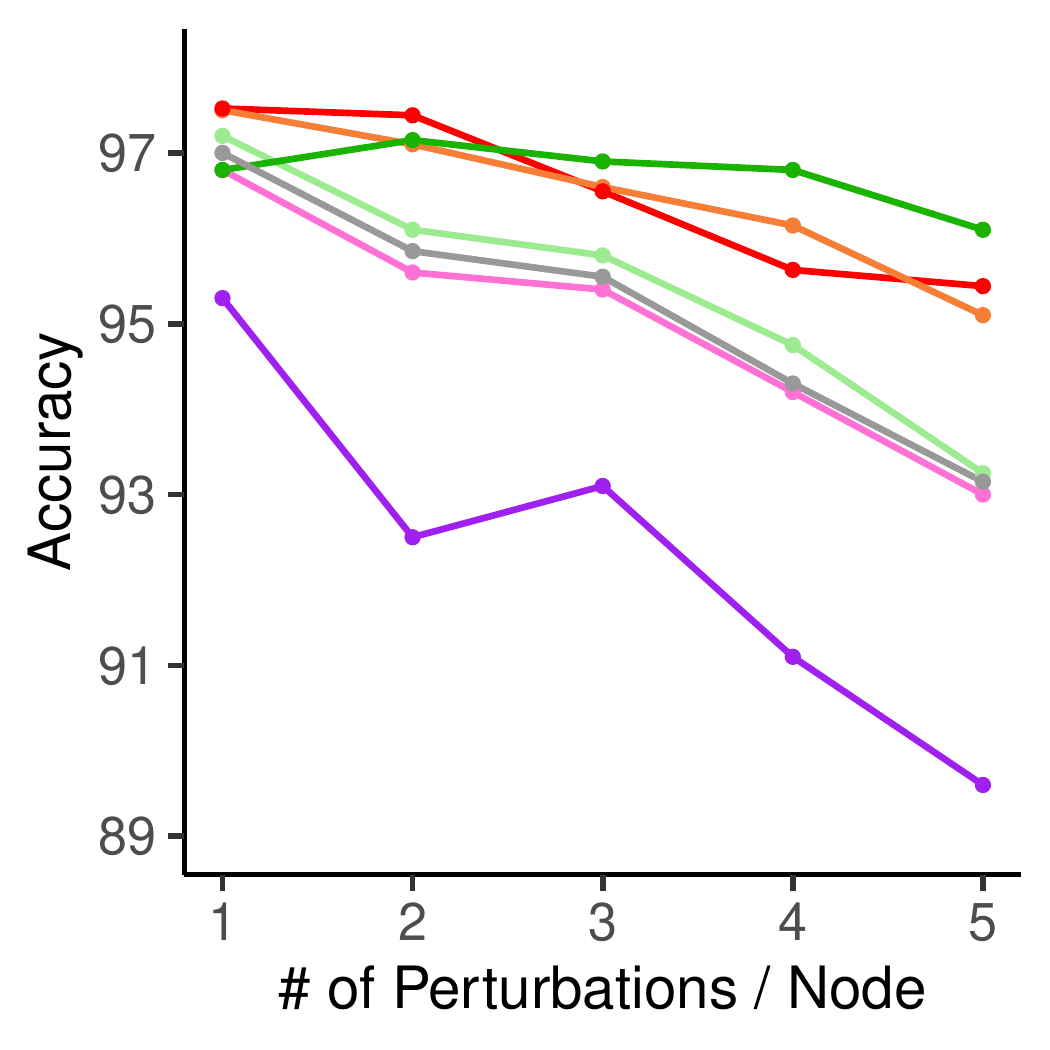}
\caption{PolBlogs}
\end{subfigure}
	\caption{Mean classification accuracy on Cora (left part), CiteSeer (middle part) and PolBlogs (right part) under targeted training-time attack (netattack) with different number of perturbations on the target nodes: (1, 2, 3, 4, 5).}
\label{fig:netattack}
\end{figure*}
%%%%%%%%%%%%%%%%

%%%%%%%%%%%%%%%%
\begin{figure*}[t!]
\centering
\includegraphics[width=0.69\textwidth]{Figures/Legend_V2.pdf}

\begin{subfigure}{0.28\textwidth}
\centering
\includegraphics[width = \textwidth, height=1.7in]{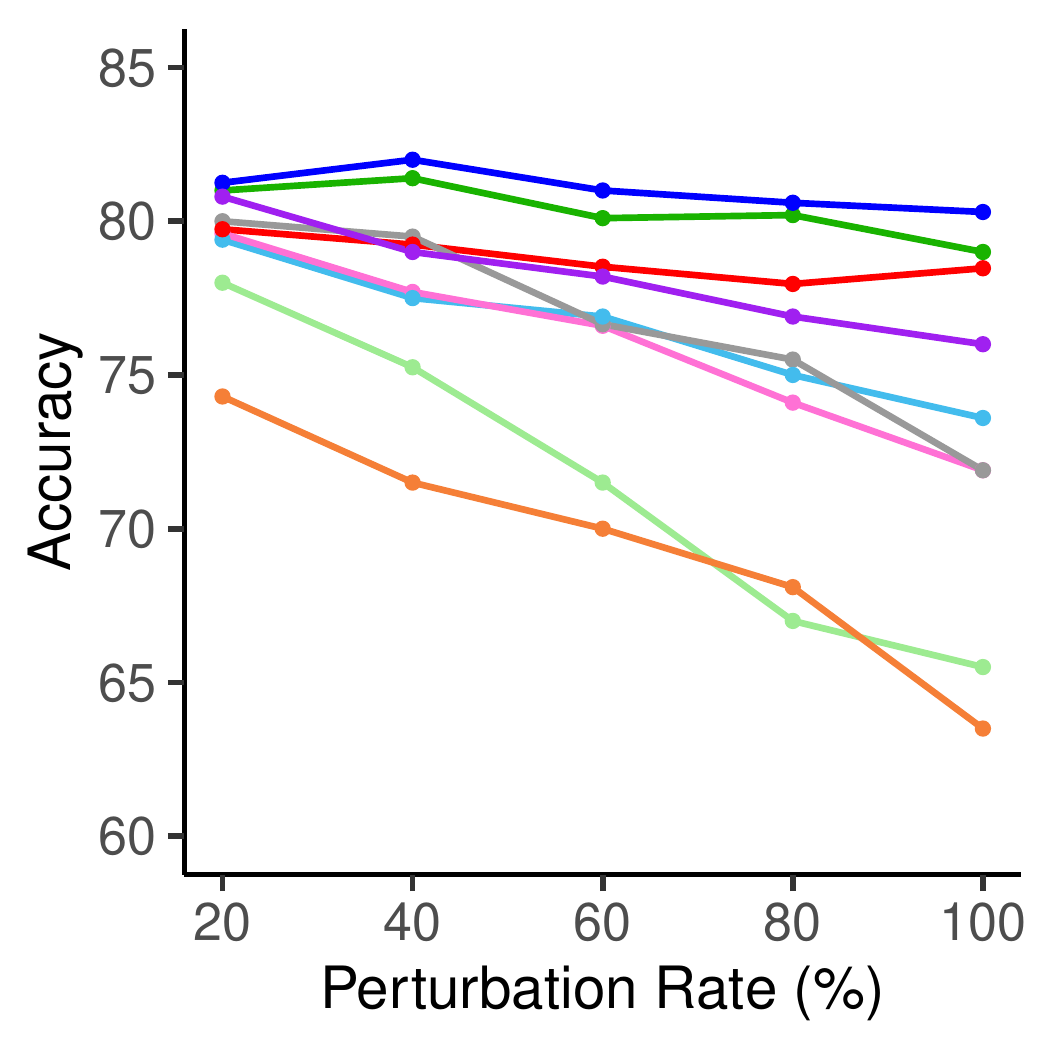} 
\caption{Cora}
\end{subfigure}
\begin{subfigure}{0.28\textwidth}
\centering
\includegraphics[width = \textwidth, height=1.7in]{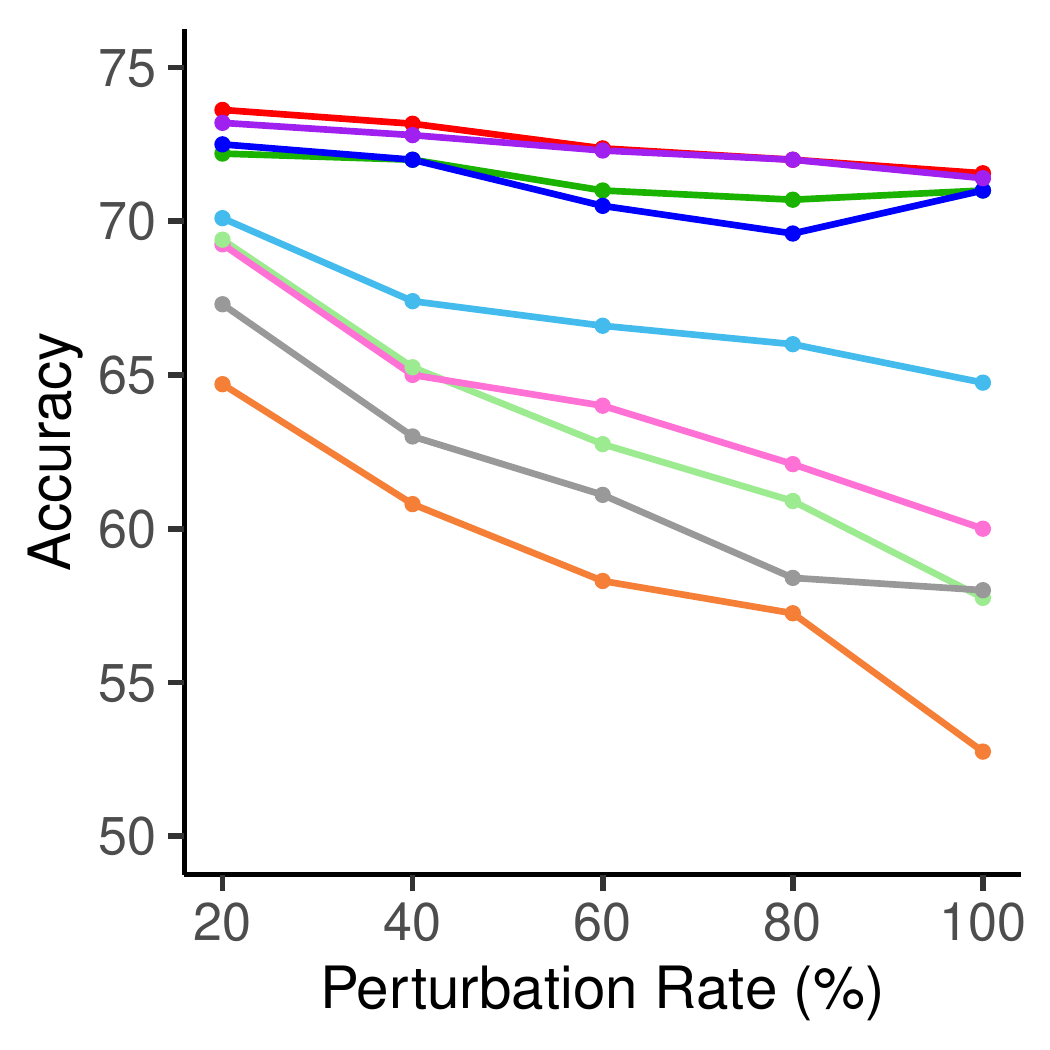}
\caption{CiteSeer}
\end{subfigure}
\begin{subfigure}{0.28\textwidth}
\centering
\includegraphics[width = \textwidth, height=1.7in]{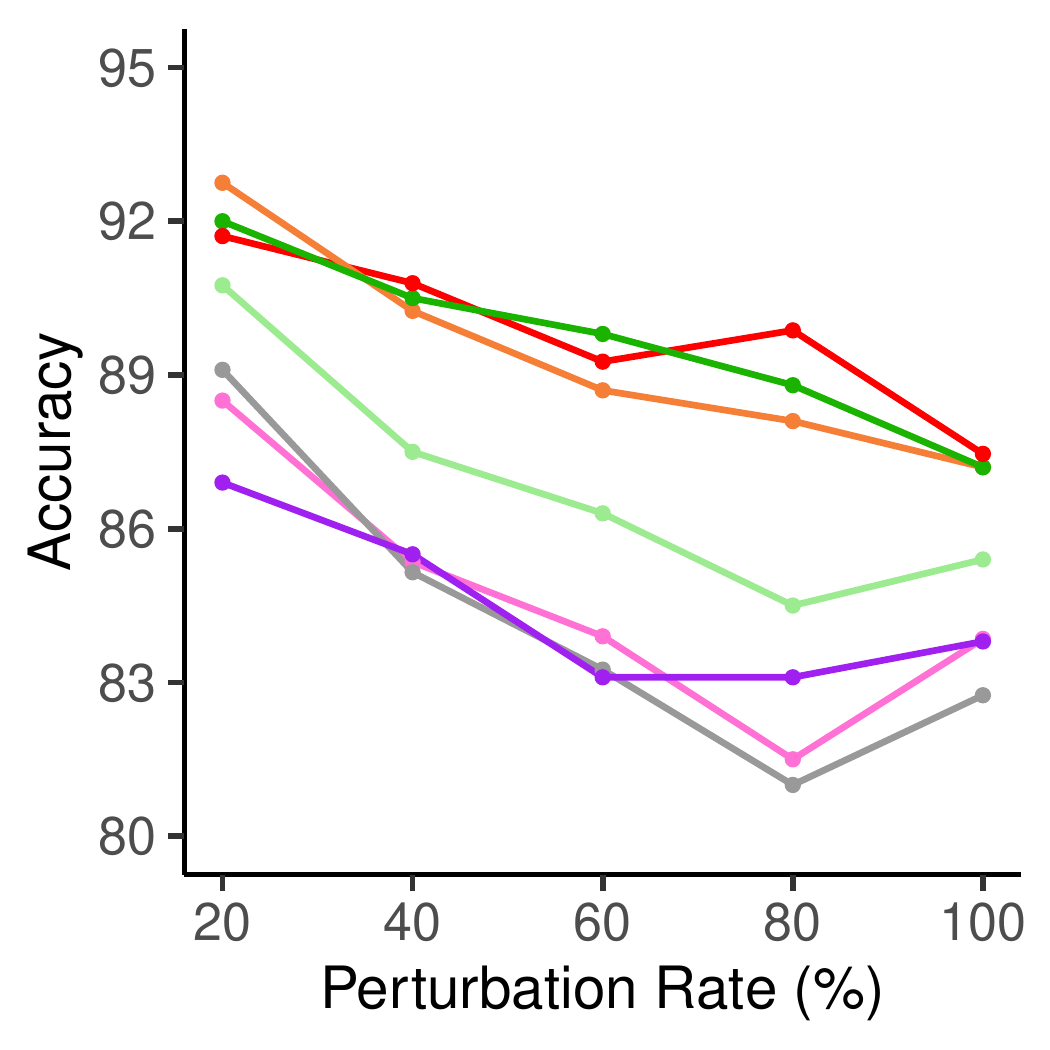}
\caption{PolBlogs}
\end{subfigure}
\caption{Mean classification accuracy on Cora (left part), CiteSeer (middle part) and PolBlogs (right part) under random training-time attack with different perturbation rates (20\%, 40\%, 60\%, 80\%, 100\%).}
\label{fig:randattack}
\end{figure*}
%%%%%%%%%%%%%%%%

\subsubsection{Implementation Details}

For GCN, GAT, RGCN, GCN-Jaccard, GCN-SVD, Pro-GNN-fs and Pro-GNN,  
we use the reported results in \cite{jin2020graph}. 
For E-GNN, we use the reported results in \cite{liu2021elastic}. 
As for our proposed ELR-GNN, 
GCN is adopted as the GNN model, which is made up of two message passing layers with Relu as the activation function.
We trained our proposed model for 1000 epochs, 
utilizing an Adam optimizer for the GNN model with a learning rate of 1e-2 and weight decay of 5e-4 
and employing a Stochastic Gradient Descent (SGD) optimizer for learning $U_d$ with a momentum of 0.9. 
The learning rate of the SGD optimizer and the values of $d$,
$\epsilon$, $\lambda_{sim}$ and $\lambda_{Fr}$ are all determined using cross-validation.
For each experiment, we report the mean and standard deviation across five runs.

\subsection{Defense Against Global Attacks}

We first evaluate the robustness of our proposed method 
against the non-targeted global attacks, which aim to degrade the overall performance of the attacked GNNs. 
For this purpose we adopt the mettack global attack \cite{zugner2019meta} with perturbation rates of $\{0, 0.05, 0.10, 0.15, 0.20, 0.25\}$. We employ the same parameters for mettack as \cite{jin2020graph} for fair comparison 
where the Meta-Self variant of mettack is used, which is one of the most challenging attack variant to defend against. 

Under this setup, we compare our proposed method, ELR-GNN, with the baselines mentioned in the previous section. 
On each dataset, 
the adjacency matrix is poisoned using mettack first, then ELR-GNN and the baselines are trained using the poisoned adjacency matrix. The classification results on the test nodes are reported in Table \ref{table:meta} where the top part of the table shows the results on the Cora dataset, the middle part shows the results on the CiteSeer dataset, and the bottom part shows the results on the PolBlogs dataset. 

We can observe that the performance of all the comparison methods degrades in general as the perturbation rate increases. 
GCN has the worst performance degradation as it depends on the graph structure to perform message propagation 
with no defense or filtering mechanisms 
against adversarial attacks on the graph structure. 
GAT performs better than GCN and RGCN on Cora and CiteSeer as its self-attention mechanism helps in learning importance weights for the edges during message passing. 
On PolBlogs, where there are no node features, GAT has large performance drops 
due to its self-attention mechanism's dependence on node features. 
Although GCN-SVD and GCN-Jaccard outperform GCN, GAT and RGCN, 
they perform much worse than the remaining defense methods on the citation datasets. 
This suggests these preprocessing based defense methods lack sufficient capacity in defending against 
state-of-the-art global attacks. 
In contrast, both ELR-GNN and the Pro-GNN variants learn the graph structure jointly with the GNN model, 
which enables these methods to defend effectively against such complex attacks.

The proposed ELR-GNN method outperforms all the other baselines on the CiteSeer datasets across almost all perturbation rates, with greater performance improvements at larger perturbation rates. 
In particular, at a 25\% perturbation rate, ELR-GNN outperforms GCN by over 16\% and 
outperforms the second-best performing baseline (Pro-GNN) by over 4\%. 
On the Cora dataset, a similar pattern of ELR-GNN outperforming all other baselines at high perturbation rates 
persists, where ELR-GNN yields even larger performance improvements of around 29\% and 6\% 
compared to GCN and the closest-performing baseline (SimP-GCN), respectively, at the 25\% perturbation rate. 
On the PolBlogs dataset, our proposed method substantially outperforms all the other baselines 
with large margins at the high perturbation rate of 25\%, yielding a particularly remarkable increase in 
test accuracy (over 13\%) over the second-best baseline (Pro-GNN-fs).

%%%%%%%%%%%%%%%%
\begin{table*}[!t] 
\caption{Mean classification accuracy (Acc), training time (Tr-T) and total time (Tot-T) of low-rank defense methods 
on Cora (top part), CiteSeer (middle part)
and PolBlogs (bottom part)
under non-targeted 
	global training-time attack (mettack) with the given perturbation rate (25\%).}
\setlength{\tabcolsep}{8pt}
\centering
\resizebox{.85\textwidth}{!}
	{
\begin{tabular}{l|lll|lll|lll}
\hline	
         & \multicolumn{3}{l|}{GCN-SVD}                   & \multicolumn{3}{l|}{Pro-GNN}                   & \multicolumn{3}{l}{ELR-GNN}                         \\
         \hline
         & Acc & Tr-T (s) & Tot-T (s) & Acc & Tr-T (s) & Tot-T (s) & Acc       & Tr-T (s) & Tot-T (s) \\
\hline
Cora     & 49.7    & 58.9             & 65.0          & 69.7    &     67,049.5  &  	67,049.5      & \textbf{76.7} & 112.1            & 118.6         \\
\hline
Citeseer & 64.8    & 58.2             & 62.4          & 68.9    & 43,894.6         & 43,894.6      & \textbf{73.2} & 85.4             & 89.4          \\
\hline
PolBlogs & 52.0    & 21.6             & 22.4          & 63.1    & 7,602.0          & 7,602.0       & \textbf{76.7} & 51.1             & 52.0         \\
\hline

\end{tabular}
}
\label{table:Time}
\end{table*}
%%%%%%%%%%%%%%%%

\subsection{Defense Against Targeted Attacks}

In this section, we investigate the robustness of the proposed ELR-GNN against 
the targeted poisoning training-time attacks on the graph structure, which target specific nodes with the aim of deceiving GNNs into misclassifying the target nodes. 

For this purpose, we employ nettack \cite{zugner2018adversarial} with different numbers of perturbations 
allowed for each target node, ranging from $\{1, 2, 3, 4, 5\}$ perturbations per target node. 
Nettack perturbs the graph structure for the target nodes by iteratively generating sets of candidate perturbations 
and applying the perturbation that would degrade the performance of a surrogate GNN model the most. 
This process is repeated until the perturbation budget has been reached. 
The target test nodes are selected similarly to \cite{jin2020graph} for a fair comparison. 

As presented in Figure \ref{fig:netattack}, 
the performance of all the comparison methods degrades as the number of perturbations on the target nodes increases. 
The proposed ELR-GNN outperforms all the other methods on Cora and CiteSeer with larger perturbation numbers,
and the improvement in performance increases as the number of perturbations per node grows,
achieving a notable performance increase of around 4\% and 6\% over the second best method on Cora and CiteSeer, respectively,
with 5 perturbations per target node. 
SimP-GCN and the two Pro-GNN variants are among the second best performing methods after our ELR-GNN, 
significantly outperforming all the other methods on the citation datasets. 
This validates the ability of the proposed ELR-GNN in defending against sophisticated targeted adversarial attacks 
such as nettack. 
On the PolBlogs dataset, ELR-GNN, Pro-GNN and GCN-SVD obtain similar results 
while the other methods perform poorly. 
These results demonstrate the robustness of our ELR-GNN against targeted training-time attacks.

\subsection{Defense Against Random Attacks}

In this section, we evaluate the robustness of our proposed ELR-GNN under random attacks that inject random edges in the graph structure with different edge perturbation rates: 20\%, 40\%, 60\%, 80\%, and 100\%. 
According to the results reported in Figure \ref{fig:randattack}, 
ELR-GNN, SimP-GCN and the two variants of Pro-GNN obtain relatively stable performance across all perturbation rates 
on Cora and CiteSeer. 
On the other hand, the performance of the other methods degrades significantly as the perturbation rate increases. 
ELR-GNN produces the best results on CiteSeer across all the perturbation rates. 
On the Cora dataset, ELR-GNN obtains a performance that is within 1-2\% of the best performing Pro-GNN
across all perturbation rates. 
On PolBlogs, ELR-GNN, Pro-GNN and GCN-SVD obtain similar results across all the perturbation rates 
and significantly outperform the remaining methods. 
Among the three methods, 
ELR-GNN slightly outperforms Pro-GNN and GCN-SVD at three out of the five perturbation rates.

%%%%%%%%%%%%%%%%
\begin{table*}[!t] %\normalsize
\centering
\caption{Ablation study results in terms of mean classification accuracy (standard deviation is within brackets) on PolBlogs
	under non-targeted training-time attack (mettack) with different perturbation rates (0\%, 5\%, 10\%, 15\%, 20\%, 25\%).}
\resizebox{.75\textwidth}{!}{
\begin{tabular}{l|l|l|l|l|l|l}
\hline	

 & 0\%                   & 5\%                   & 10\%                  & 15\%                  & 20\%                  & 25\%                  \\
    \hline
GCN                 & $95.6_{(0.3)}$          & $73.0_{(0.8)}$          & $70.7_{(1.1)}$          & $64.9_{(1.9)}$          & $51.2_{(1.2)}$          & $49.2_{(1.3)}$          \\
\hline
ELR-GNN     & $\mathbf{95.9}_{(0.1)}$ & $\mathbf{94.5}_{(0.2)}$ & $88.2_{(0.1)}$          & $\mathbf{80.4}_{(0.5)}$ & $\mathbf{77.5}_{(2.0)}$ & $\mathbf{76.7}_{(1.2)}$ \\
$\quad - \text{w/o } \mathcal{L}_{Sim}$ & $95.8_{(0.2)}$          & $93.7_{(0.8)}$          & $88.1_{(0.2)}$          & $53.5_{(3.3)}$         & $67.5_{(2.2)}$          & $52.2_{(0.8)}$          \\
$\quad - \text{w/o }  \mathcal{L}_{Fr}$   & $95.8_{(0.0)}$          & $94.1_{(1.0)}$          & $88.1_{(0.1)}$         & $75.4_{(1.1)}$          & $66.7_{(5.0)}$          & $67.1_{(5.9)}$          \\
$\quad - \, \epsilon = 0$ & $95.7_{(0.1)}$          & $93.4_{(0.5)}$          & $87.8_{(0.3)}$          & $76.8_{(2.9)}$          & $75.6_{(2.8)}$          & $75.9_{(2.3)}$          \\
$\quad - \, \text{Rand. Init.} $  & $50.5_{(1.4)}$          & $51.6_{(1.1)}$          & $51.3_{(1.9)}$         & $74.0_{(4.3)}$          & $49.2_{(1.7)}$          & $50.2_{(2.3)}$          \\
$\quad - \,\text{Joint Update}$      & $95.6_{(0.1)}$          & $93.2_{(1.1)}$   & $\mathbf{88.4}_{(0.1)}$ & $76.9_{(3.4)}$          & $66.7_{(5.1)}$          & $72.9_{(5.2)}$      \\   
\hline
\end{tabular}
\label{table:ablation}
} 

\end{table*}
%%%%%%%%%%%%%%%%

\subsection{Efficiency Analysis}

In order to demonstrate the efficiency of the proposed ELR-GNN, we summarize the training time, total time (pre-processing time + training time) and accuracy of all the three low-rank based defense methods: ELR-GNN, Pro-GNN and GCN-SVD. 
We conduct experiments on Cora, CiteSeer and PolBlogs under global training-time attacks (mettack) with a perturbation rate of 25\%. We run all the experiments on NVIDIA GeForce RTX 2080 ti and train each method for 1000 epochs. 
For Pro-GNN, we use the optimal hyper-parameters reported in \cite{jin2020graph}. 
For GCN-SVD, we use the same $d$ value employed for ELR-GNN on each dataset for a fair comparison. 

The corresponding results are reported in Table \ref{table:Time}. 
It is clear that ELR-GNN not only outperforms Pro-GNN in terms of classification accuracy 
but is also significantly more efficient. 
ELR-GNN is 560, 500 and 150 times faster than Pro-GNN on Cora, CiteSeer and PolBlogs, respectively. 
Pro-GNN requires long training times mainly due to optimizing the nuclear norm of the learned adjacency matrix in each iteration. 
Additionally, ELR-GNN significantly outperforms 
GCN-SVD by 26\%, 9\% and 24\% on Cora, CiteSeer and PolBlogs respectively
in terms of classification accuracy,
while being only a bit more than 2 times slower or less. 
This clearly demonstrates the efficiency and robustness of ELR-GNN in defending GNNs against global adversarial structural attacks.

\subsection{Ablation Study}

We conduct an ablation study to investigate the effect of each component of the proposed ELR-GNN. 
Specifically, we examine the following variants of ELR-GNN: 
(1) ``$\text{w/o } \mathcal{L}_{Sim}$", where the similarity regularization term $\mathcal{L}_{Sim}$ is dropped. 
(2) ``$\text{w/o } \mathcal{L}_{Fr}$", where the Frobenius norm regularization term $\mathcal{L}_{Fr}$ is dropped. 
(3) ``$\epsilon = 0$", where the sparsification is dropped by setting the adjacency matrix sparsity threshold to zero. 
(4) ``$\text{Rand. Init.}$", where $U_d$ is initialized randomly using 
the Xavier normal initialization instead of the truncated SVD.
(5) ``Joint Update" variant, where the GNN model and $U_d$ are updated simultaneously 
rather than alternately at each training iteration. 
We compare the performance of these variants with ELR-GNN and the GCN baseline, 
where the GCN baseline can be treated as the variant that drops the entire proposed defense method. 
We report the performance of GCN, ELR-GNN and all the variants on the PolBlogs dataset 
under the mettack based poisoning attacks with perturbation rates of $\{0, 0.05, 0.10, 0.15, 0.20, 0.25\}$ 
in Table \ref{table:ablation}.

From the table, it is clear that the ``$\text{Rand. Init.}$" variant performs very poorly across all the perturbation rates, which highlights the importance of using truncated SVD to initialize the singular vector matrix. 
All the other variants perform very similarly to ELR-GNN but significantly outperform GCN 
under the low-perturbation rates (0\%, 5\%, and 10\%). 
This indicates that the low-rank estimate of the adjacency matrix obtained 
using SVD is only modified in a minor fashion during the training stage in such cases, 
which is consistent with the fact that the low perturbation rates only 
lead to very limited disturbances on the graph structure. 
At the higher perturbation rates (15\%, 20\%, and 25\%), 
the performance of all the variants has notable drops from the full ELR-GNN. 
In particular, the ``$\text{w/o } \mathcal{L}_{Sim}$" variant obtains very poor results, which indicates the importance of learning a graph adjacency matrix that is not significantly different from the input adjacency matrix. 
The ``$\epsilon = 0$" variant performs only slightly worse than ELR-GNN, 
which indicates that it is possible to learn dense matrices that perform similarly to their sparse counterparts. 
However, such dense adjacency matrices would be significantly more expensive to learn relative to sparse adjacency matrices. 
The ``Joint Update" variant also demonstrates noticeable performance declines
from the full ELR-GNN,
which highlights the challenging nature of the optimization problem at high perturbation rates, 
where simultaneous updates are not as effective as alternating updates. 
Overall, the results in Table \ref{table:ablation} demonstrate the contribution of each component of 
the proposed ELR-GNN for learning robust GNN models against adversarial attacks on graph structures.

\section{Conclusion}

In this paper, we proposed a novel ELR-GNN method to defend GNNs against sophisticated adversarial attacks on the graph structure. 
The proposed framework learns a low-rank sparse estimate of the adjacency matrix as the product of low-dimensional matrices,
and is made up of two modules: a coarse low-rank estimation module and a fine-grained estimation module. 
The coarse low-rank estimation module employs the truncated SVD
to calculate the singular value matrix and initialize the low-dimensional singular vector matrix. 
Then the fine-grained estimation module learns a robust low-rank and sparse adjacency matrix 
by jointly optimizing the singular vector matrix and the GNN model. 
The weak edges in the estimated adjacency matrix are pruned to sparsify the matrix. 
We conducted comprehensive experiments under
three different training-time attacks on the graph structure. 
The experimental results demonstrated that ELR-GNN is more robust to adversarial attacks than other 
existing GNN defense methods and can be trained in an efficient manner. 
%%%%%%%%% REFERENCES

\bibliographystyle{ieeetr}
\bibliography{egbib}

\end{document}